\documentclass{esp-paper}
\title{Beyond task performance:\\ Decoding bioacoustic embeddings with speech features}
\shorttitle{Decoding bioacoustic embeddings}

\author{Ines Nolasco$^{\star}$}
\author{Jules Cauzinille}
\author{Marius Miron}
\author{Gagan Narula}
\author{Milad Alizadeh}
\author{\authorcr Emmanuel Fernandez}
\author{Matthieu Geist}
\author{Ellen Gilsenan-McMahon}
\author{Olivier Pietquin}
\author{\authorcr Emmanuel Chemla}
\author{Sara Keen$^{\ddag}$}
\affil{Earth Species Project}

\usepackage{tabularx}

\begin{document}
\thispagestyle{firstpage}
\maketitle

\begingroup
\def\thefootnote{$\star$}\footnotetext{Corresponding author: \url{ines@earthspecies.org}}
\def\thefootnote{$\ddag$}\footnotetext{Senior author}
\def\thefootnote{\arabic{footnote}}
\endgroup

\begin{abstract}
Pretrained audio embeddings are standard in bioacoustics, yet little is
known about which acoustic features these models encode, nor which are
useful for a given task. This hinders transparency and limits extension
to rare species or data-scarce domains. Here we reveal which speech-like
features are encoded in bioacoustic representations. Using the 88~eGeMAPS
features across six taxonomic groups, we apply linear and nonlinear
regression probes to quantify which acoustic properties each model
captures. Results confirm a ``no free lunch'' pattern: no single model
captures the full feature space. A concatenated embedding achieves the
highest performance, suggesting complementary acoustic space coverage across
models. Loudness features are best encoded (R\textsuperscript{2}~=~0.76)
while F0 is hardest to recover (R\textsuperscript{2}~=~0.33). By
cross-referencing recoverability with per-species feature salience (NMI),
we derive data-driven model selection guidance for bioacoustics.
\end{abstract}


\begin{figure*}[htbp]
    \centering
\includegraphics[width=.85\textwidth]{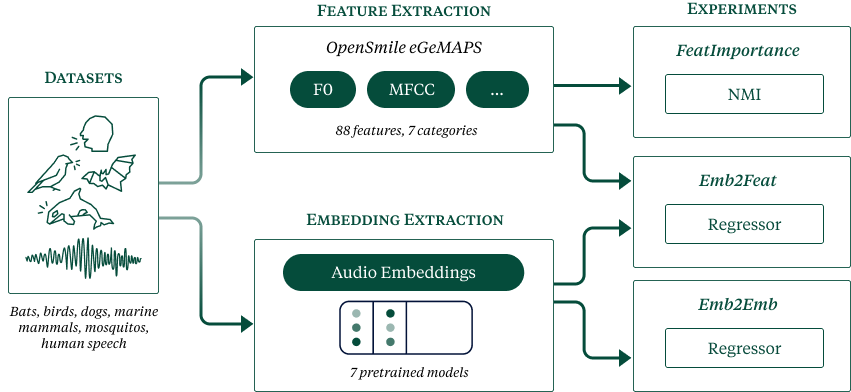}
    \caption{Starting from a variety of domains (left), we ask which acoustic features (extracted with openSMILE, top) are represented by which bioacoustic models (bottom).}
    \label{fig:overall_approach}
\end{figure*}

\section{Introduction}

Computational bioacoustics has been transformed by deep learning, enabling
large-scale acoustic monitoring and finer-grained ecological questions
\parencite{parker2025planetization}. Despite strong task performance in species
classification and detection, it remains largely unclear which acoustic
characteristics drive model decisions---interpretability is limited, and
the field is only beginning to address this systematically.
A common practice in current pipelines is the use of pretrained embeddings.
Models are trained on large, diverse corpora to learn general-purpose
representations that are then reused for downstream tasks
\parencite{stowell2022computational}. While this transfer learning paradigm has
shown strong results---especially in low-data regimes---the internal
representations are opaque. Unlike handcrafted acoustic features, which
have direct physical interpretations, it is unknown which properties of
sound are preserved, amplified, or discarded within embedding spaces, or
how they align with interpretable acoustic characteristics.
Several pretrained models now exist for non-human vocal analysis, spanning
supervised and self-supervised paradigms and trained on speech, general
audio, and bioacoustic data. Whether these models learn similar or
complementary representations remains an open question. Recent benchmarking
initiatives compare models on downstream classification performance
\parencite{miron2025matters}, but do not directly reveal \emph{what} information
is encoded. Consequently, principled guidelines for model selection are
still lacking.

In this work, we address the following research question: \emph{Which
acoustic features are encoded in the embedding representations of deep
learning models used in bioacoustics?} We extract 88~interpretable acoustic
descriptors from the eGeMAPS feature set---covering spectral, temporal,
cepstral, and modulation-based properties---from six bioacoustic domains: dogs, mosquitoes, bats, marine mammals, birds, and human speech. We then apply linear and
nonlinear regression probes to measure how well these features can be recovered from embeddings produced by pretrained models (see \parencite{belinkov2022probing}).
The performance gap between the two serves as an indicator
of representational entanglement.
To connect recoverability to practical utility, we further compute the Normalised Mutual Information (NMI) between each feature and labels of each dataset.
Cross-referencing NMI with probe $R^2$ reveals whether models encode the features that are acoustically salient for a given task, providing a principled basis for model selection beyond benchmark scores.
Our contributions are threefold:
\begin{itemize}
    \item A regression probing framework to measure feature recoverability from pretrained embeddings, enabling comparison of \emph{what} models encode rather than task performance alone.
    \item An analysis of 6 pretrained models across 88~eGeMAPS features and six bioacoustic datasets, revealing substantial cross-model variation and complementarity.
    \item A cross-referencing of feature recoverability with per-species feature salience (NMI), providing practical insights into model fitness for specific bioacoustic tasks.
\end{itemize}

\section{Background}
\label{sec:background}

Early machine listening and speech research focused on identifying
measurable signal properties such as frequency, amplitude and temporal
structure, that provide compact, interpretable representations of complex
sounds \parencite{sharma2020trends}. To standardise extraction, the openSMILE
toolkit \parencite{eyben2010opensmile} was developed, enabling reproducible
feature computation across benchmarks such as the INTERSPEECH
Computational Paralinguistics Challenge \parencite{schuller2013interspeech}.
Within this framework, the Geneva Minimalistic Acoustic Parameter Set
(GeMAPS) and its extended version (eGeMAPS) were curated as a principled,
minimal feature set for speech-related tasks, particularly emotion
recognition.
Although developed for speech, these features transfer well to bioacoustics:
there is substantial overlap in informative acoustic properties across
human and non-human vocalisations \parencite{torrisi2026embryonic,
fagerlund2007bird}, and Mel-frequency Cepstral Coefficients (MFCCs)
achieved early successes in bioacoustic classification
\parencite{deecke2006automated, somervuo2006parametric}. Nevertheless,
bioacoustic signals can differ markedly from speech. For instance, bats vocalise in the
ultrasonic range \parencite{jones2007bat}, elephants use infrasonic rumbles
\parencite{payne1986infrasonic}, and insects produce sound through stridulation
rather than vocal tracts \parencite{alexander1957sound}. Furthermore, bioacoustic recordings are
often made in natural environments, with overlapping species and varying background noise conditions. Finding
representations that generalise across taxa and deployment conditions
therefore remains challenging \parencite{stowell2018bioacoustic}.

With the rise of deep learning, the field moved away from handcrafted
features toward representations learned directly from data. Pretrained
embeddings have become the dominant paradigm: a model is trained on large
corpora and its frozen encoder representations are reused for downstream
tasks with minimal additional training \parencite{stowell2022computational}.
Success cases \parencite{ghani2023global, best2023deep} have motivated broad
adoption, but results are mixed and difficult to predict, particularly under
domain mismatch between pretraining and target data \parencite{miron2025matters}.

Benchmarking efforts to guide model selection have been established in both speech SUPERB
\parencite{yang2021superb}, HEAR \parencite{turian2022hear} and, more recently,
bioacoustics \parencite{miron2025matters, kather2025clustering,
schwinger2025foundation}. These evaluations typically extract embeddings
from frozen encoders and assess them with a simple classification head. The
choice of classification head is key: shallow multilayer perceptrons (MLPs) allow some non-linear adaptation of
the embedding space \parencite{turian2022hear}, while linear probes
\parencite{belinkov2022probing} impose stricter constraints. Clustering or
retrieval methods require no training at all, directly reflecting the
geometry of the embedding space \parencite{miron2025matters, kather2025clustering}.
This shift from handcrafted to learned features has come at the cost of
interpretability. Understanding what acoustic patterns models exploit is
important both for transparency in applied conservation contexts
\parencite{parker2025planetization} and for scientific discovery, since learnt
representations can reveal biologically meaningful acoustic properties
relevant to animal communication and evolution studies.
In speech, probing studies have established that self-supervised models
such as wav2vec~2.0 encode information hierarchically across layers: shallow
layers capture acoustic properties such as fundamental frequency and
formants, middle layers encode phonetic content, and deeper layers represent
word identity and semantics \parencite{pasad2021layer, pasad2023comparative, choi2022opening}. A clear
layer-wise correspondence with MFCCs has also been demonstrated
\parencite{raymondaud2024probing}.

In bioacoustics, this research direction is still underdeveloped. An important
insight from \textcite{kather2025clustering} links training paradigm to
representation content, finding that supervised models tend to produce better-structured clusters, supporting species classification, while self-supervised models encode more general characteristics of the sounds and thus allow better generalization across domains.
Beyond this, it remains unclear what different models actually encode, whether their representations are complementary or redundant, and, critically, whether they capture the features that are most discriminative for particular taxonomic groups.
On the whole, our current understanding of bioacoustic embeddings lags far behind that of speech encoding. Addressing this gap is the central aim of the present work.

\section{Materials and Methods}
\label{sec:matandmet}

Fig.~\ref{fig:overall_approach} depicts our methodological framework. We use six bioacoustic audio datasets spanning diverse taxa (\S\ref{sec:data}). For each recording, we extract a set of interpretable acoustic features (\S\ref{sec:data}), as well as embeddings from six pretrained models (\S\ref{sec:models}).  We then conduct three complementary experiments: 1) apply regression probes to determine how recoverable features are from each embedding space (Emb2Feat); 2) apply cross-model probes to determine overlap in what information is encoded by each model (Emb2Emb); and 3) calculate importance of interpretable features and compare this against feature recoverability, giving insight into how well \emph{relevant} features have been encoded in each embedding (FeatImportance).

\subsection{Dataset and feature extraction with openSMILE}
\label{sec:data}

Our data
is extracted from the training split of the BEANS benchmark dataset described and available in \textcite{hagiwara2023beans}.
The selected sets support a range of tasks from species to individual classification (see Table~\ref{tab:datasets}).

\begin{table*}[htbp]
\centering
\caption{Dataset summary}
\label{tab:datasets}
\begin{tabularx}{\textwidth}{@{}l r X r@{}}
\toprule
\textbf{Dataset} & \textbf{\# Audio} & \textbf{Task} & \textbf{\# Class}\\
\midrule
Dogs                & 414   & IndividualID & 10  \\
Bats                & 5987 & IndividualID & 10  \\
CBI -- birds        & 14206 & Species       & 264 \\
MM -- marine mammals & 1004 & Species       & 31  \\
Mosquitoes          & 5407 & Species       & 14  \\
Speech  & 7036 & Spoken words &  35\\
\midrule
\textbf{Total}      & \textbf{34054} & - & - \\
\bottomrule
\end{tabularx}
\end{table*}

Interpretable acoustic features were extracted with openSMILE \parencite{eyben2010opensmile} (\S\ref{sec:background}). We used the eGeMAPS feature set which defines 88 global acoustic features that are further categorised in seven broader types (described in Table~\ref{tab:feature_groups}).

\begin{table*}[ht]
\centering
\small
\caption{eGeMAPSv02 feature groups used in this study (88 features total).}
\label{tab:feature_groups}
\renewcommand{\arraystretch}{1.2}
\newcolumntype{Y}{>{\small\raggedright\arraybackslash}X}
\begin{tabularx}{\textwidth}{@{} l Y r @{}}
\toprule
\textbf{Group} & \textbf{Description} & \textbf{\# Feat.} \\
\midrule
\textbf{F0} & Fundamental frequency level, range and slope & 10 \\ \addlinespace[0.5ex]
\textbf{Loud}ness & Perceptual loudness level, range  and long-term energy & 11 \\ \addlinespace[0.5ex]
\textbf{Harm}onicity & Jitter, shimmer, harmonics-to-noise ratio, harmonic amplitude ratios & 10 \\ \addlinespace[0.5ex]
\textbf{Sp}ectral \textbf{Sh}ape & Spectral tilt (alpha ratio, Hammarberg index), spectral slopes, spectral flux & 17 \\ \addlinespace[0.5ex]
\textbf{Form}ants & F1, F2, F3 frequency, bandwidth, and amplitude (voiced frames) & 18 \\ \addlinespace[0.5ex]
\textbf{MFCC} & Mel-frequency cepstral coefficients 1--4 (all frames and voiced frames) & 16 \\ \addlinespace[0.5ex]
\textbf{Temp}oral & Voiced/unvoiced segment rate and duration, loudness peak rate & 6 \\
\bottomrule
\end{tabularx}
\end{table*}

\subsection{Models}
\label{sec:models}
We extract pre-trained embeddings from the \textit{last layer} of six different models that are commonly used in bioacoustics research and have been the subject of recent evaluation in \textcite{miron2025matters}. Models were pretrained on various domains and with different learning paradigms (see Table~\ref{tab:models}). Extraction and access to the models' checkpoints is done with the AVEX API\footnote{\url{https://github.com/earthspecies/avex}}.
Embeddings are averaged on the time dimension for the whole clip resulting in a single vector per sample.

\begin{table*}[ht]
\centering
\small
\caption{Summary of Audio encoder models.}
\label{tab:models}
\renewcommand{\arraystretch}{1.2}
\begin{tabularx}{\textwidth}{@{} l l l l r @{}}
\toprule
\textbf{Model} & \textbf{Archit.} & \textbf{Learn.} & \textbf{Training Domain} & \textbf{Dim} \\
\midrule
BEATS\_base \parencite{chen2022beats}         & Transf.   & SSL   & general audio      & 768  \\ \addlinespace[0.5ex]
NatureLM \parencite{robinson2024naturelm} & Transf.   & SSL + LLM  & Bio + Speech + Music           & 768  \\ \addlinespace[0.5ex]
\addlinespace[0.5ex]
BirdMAE \parencite{rauch2025birdmae}      & Transf.    & SSL   & Bio           & 1280 \\ \addlinespace[0.5ex]
BirdNET \parencite{kahl2021birdnet}       & CNN  & Sup  & Bio(Birds)      & 1024 \\ \addlinespace[0.5ex]
EffNet\_all \parencite{miron2025matters}        & CNN  & Sup  & Bio + general audio  & 1280 \\ \addlinespace[0.5ex]
Perch  \parencite{ghani2023global}       & CNN  & Sup  &  Bio(Birds)  & 1280 \\

\bottomrule
\end{tabularx}
\end{table*}

\subsection{Linear and non-linear regression probes}

We define two regression probes. For the linear probe, we fit a ridge regression model using \textit{scikit-learn}\footnote{\url{https://scikit-learn.org/stable/modules/generated/sklearn.linear_model.Ridge.html}} that learns a linear mapping from the embeddings vector to each feature value.
The non-linear probe is implemented as a shallow multi-layer perceptron consisting of a single hidden layer with 256 units, ReLU activation, and dropout ($p = 0.2$), followed by a linear output unit. The network is optimised with Adam ($lr = 0.001$) using MSE loss, batch size 128 and early stopping with patience 20 (maximum 100 epochs).
For both probes, input embeddings and target features are Z-score normalised.
Predictions are inverse-transformed to the original scale before computing the evaluation metrics.
We evaluate regression performance under 5-fold cross-validation stratified by dataset and report per-feature coefficients of determination $R^2$ averaged across folds.

\subsection{Determining feature relevance for tasks with NMI}

To measure relevance of interpretable acoustic features, we computed normalized mutual information (NMI) for each dataset. This quantifies how well samples with different labels could be distinguished using each feature (higher NMI indicates higher feature importance for the task).
$\text{NMI}(X; Y) = \frac{I(X; Y)}{H(Y)}$
where $I(X; Y)$ is the mutual information between a continuous acoustic feature $X$ and the discrete class label $Y$, and $H(Y) = -\sum_c p(c) \log p(c)$ is the Shannon entropy of the label distribution (in nats). MI is estimated with a $k$-nearest neighbour entropy estimator ($k = 3$).

\section{Results}

\subsection{Emb2Feat: Recoverability of features}

\begin{figure*}[htbp]
    \centering
    \includegraphics[width=\textwidth,
    trim=0 0 0 8.5mm,clip
        ]{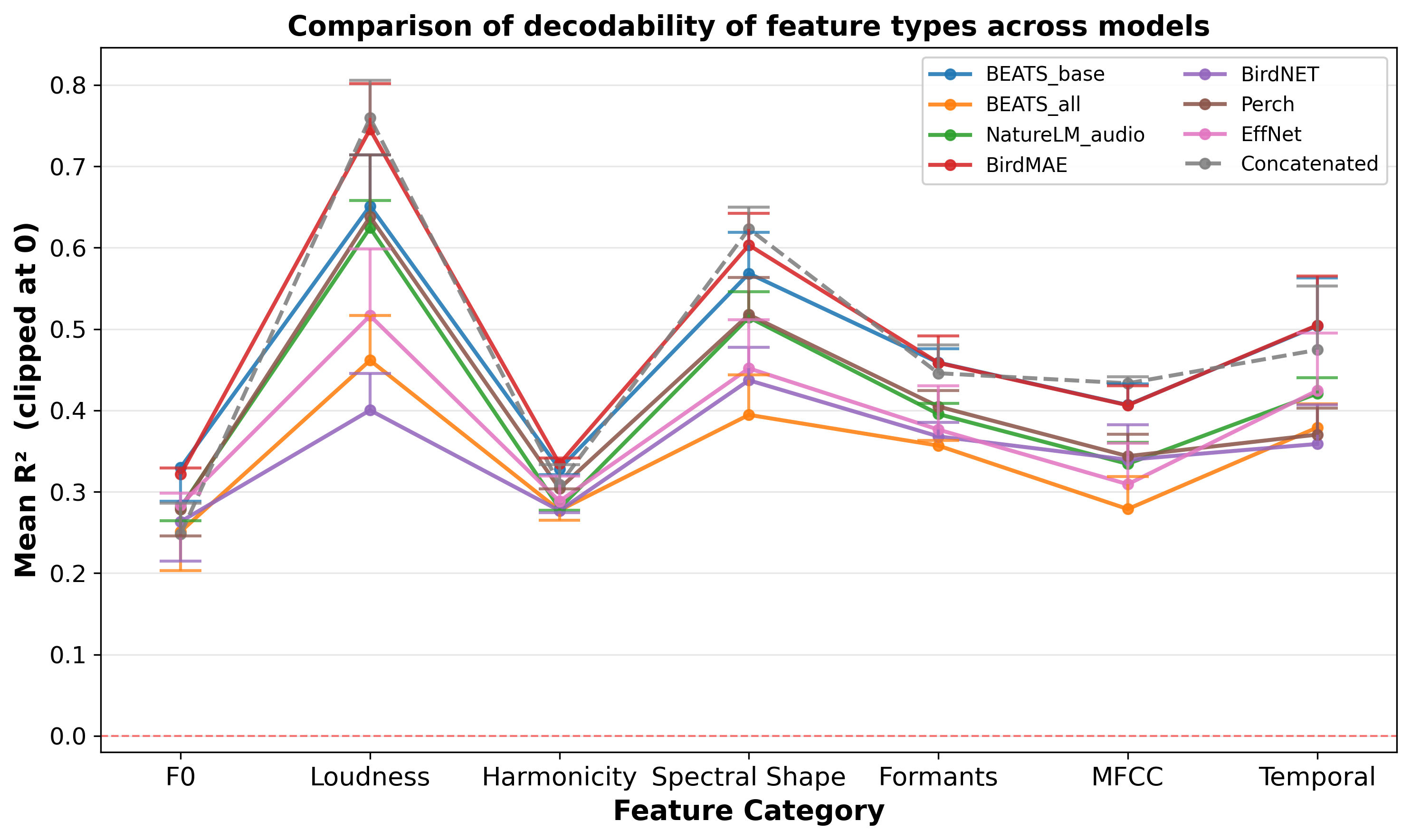}
    \caption{Mean R² scores for linear decoding of eGeMAPS feature categories (dots). Vertical stems show non-linear gain and the stem cap marks the MLP R² value. Each line corresponds to one of six pretrained audio embedding models and all embeddings concatenated. X-axis shows feature categories.
    Higher R² values indicate greater decodability of feature category from the embedding space.}
    \label{fig:r2featuretypes}
\end{figure*}

\begin{figure*}[htp]
    \centering

    \includegraphics[
        width=\textwidth,
    ]{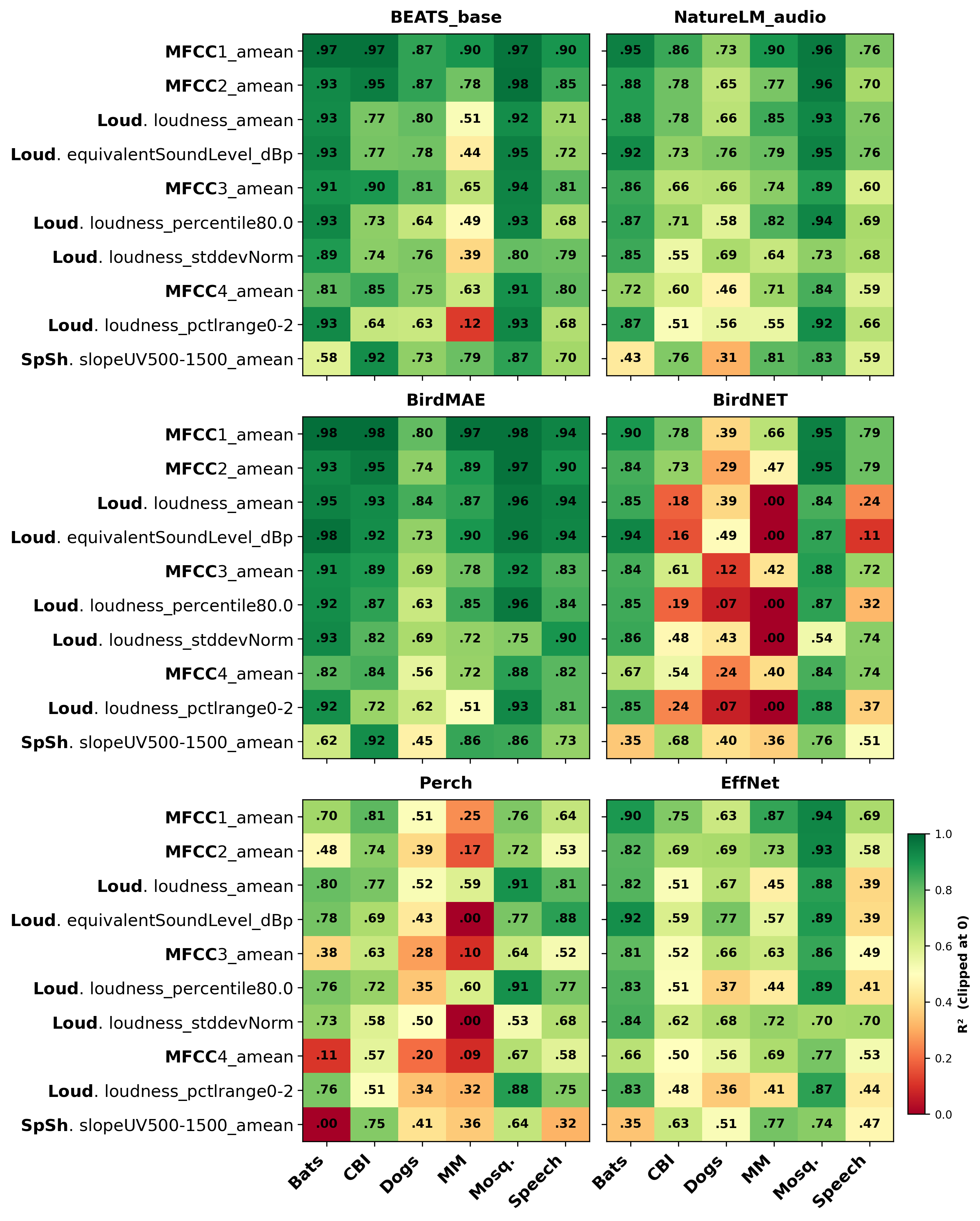}
 \caption{Union of the top5 linearly decodable eGeMAPS features for each model, shown across datasets (rows) and models (columns). Colour represents mean R² for each feature-dataset-model combination, illustrating which acoustic characteristics are most recoverable from each embedding.}
    \label{fig:recoverableFeatures_perModel}
\end{figure*}

Results of recoverability (with linear and non-linear probes) of feature types for different models are presented in Fig.~\ref{fig:r2featuretypes}.
Overall, we observe that \textbf{BirdMAE} and \textbf{BEATS\_base} embeddings are the best encoders across all feature types. This tendency might be explained by the ability of SSL models to encode general acoustic properties as suggested in \parencite{kather2025clustering}. However, due to the differences in training domain and architectures this warrants further investigation.
For most of the feature types, \textbf{concatenation} of all models (dashed line) provides the best representation; this suggests that models may be complementary in what they encode.
The general small improvement in $R^{2}$ (max +0.08) when applying a non-linear setup might indicate that the non-linearity applied is still very shallow for the level of entanglement that these embeddings present. Given these results, we proceed considering only the linear case.
Regarding feature types, loudness- and spectral shape-related features are recovered best across all models while F0-related features are worst.

Zooming in on individual features, Fig.~\ref{fig:recoverableFeatures_perModel} presents the $R^{2}$ values across models and datasets for the union of the top 5 linearly recoverable features across all models.
From here we can first observe the high variation in feature recoverability across models, but also across datasets within the same model (\textit{e.g.,} \textbf{BirdNET} and \textbf{Perch}).
It is also of note that several MFCC-related features are present, but not a single F0.
The results on the concatenated model and the overall heterogeneity observed leads us to analyse how embeddings relate to each other and in particular whether some models might be able to directly predict other models' embeddings.

\subsection{Emb2Emb}
\label{emb2emb}

\begin{figure*}[htbp]
    \centering
    \includegraphics[width=.75\textwidth]{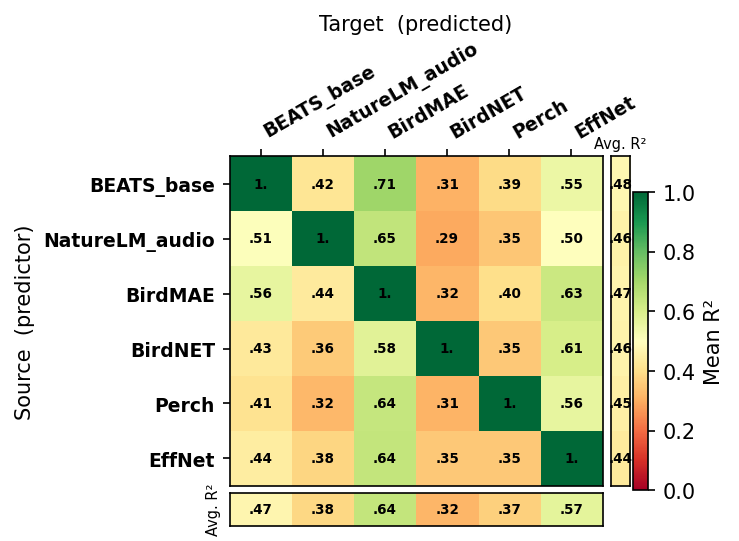}
    \caption{Pairwise embedding predictability across all models. Cells show mean R² for predicting one embedding (target, columns) from another (source, rows), using a linear probe trained on combined bioacoustic and speech datasets. Higher values indicate greater overlap in encoded information between model pairs.}
    \label{fig:pairwise_embeddings}
\end{figure*}

Fig.~\ref{fig:pairwise_embeddings} shows that no single model is best suited to predict others, although \textbf{BirdMAE} and \textbf{EffNet\_all} seem to be partially predictable by most. In contrast, \textbf{BirdNET} is the least predictable model. These results indicate that each pre-training dataset, objective, and architecture generate rather unique feature extraction capabilities, warranting further investigation, as we explore in the following section.

\subsection{FeatImportance}

\begin{figure*}[htbp]
    \centering

    \includegraphics[width=\textwidth
    ,
    height=.8\textheight,
    keepaspectratio,
     trim=0 0mm 0 20mm,
    ]{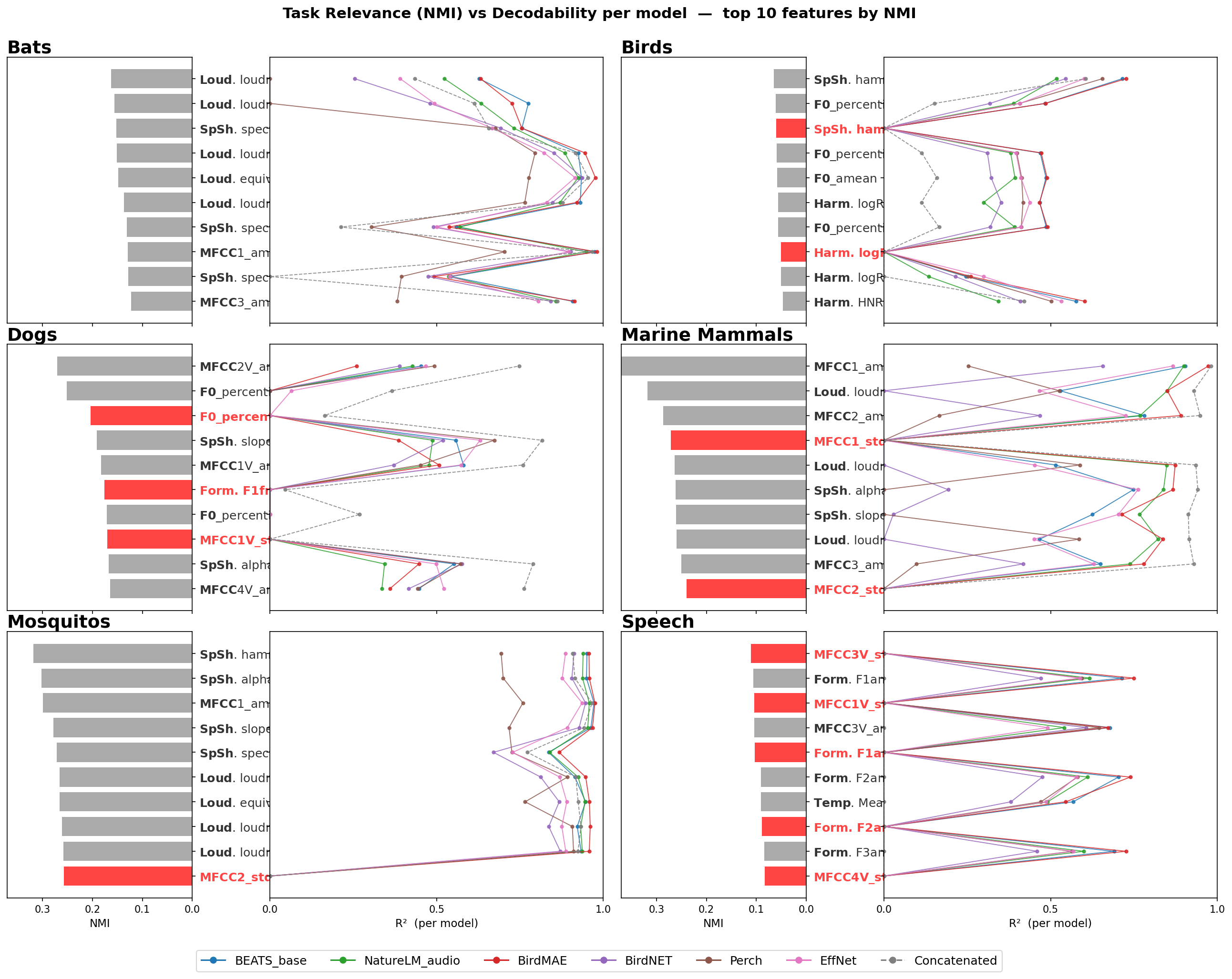}

    \caption{Task relevance versus embedding decodability of the top10 eGeMAPS features per dataset. The x-axis shows NMI; the y-axis shows mean $R^{2}$ from linear probing. Grey bars indicate features that are both task-relevant and linearly decodable; red bars indicate task-relevant features with low linearly decodability, highlighting gaps in embedding representations.}
    \label{fig:NMIxRidge}
\end{figure*}

In Fig.~\ref{fig:NMIxRidge}, we cross-reference the most acoustically salient features per task (measured by NMI) with each model's ability to encode them (measured by R\textsuperscript{2}). Salient features vary considerably across taxonomic groups: birds and dogs rely more heavily on F0-related features; mosquitoes and bats on loudness; and
marine mammals and speech commands on MFCCs.
Consistent with earlier results, loudness features are generally well-encoded while F0 and
several MFCC features remain hard to recover linearly from the embeddings of the last layer. This supports the observation that half of the top-10 salient features for speech commands
are not reliably encoded by any individual model.
Notably, concatenated embeddings do not always result in the highest recoverability of task-relevant features (\textit{e.g.,} Birds, Bats and Speech); these results might in part be explained by the high dimension of concatenated embeddings, which can easily overfit and degrade regression performance.

\section{Final remarks}
We propose an evaluation framework of embedding representations based on acoustic feature content. Our results confirmed that no single model captures the full breadth of the eGeMAPS features, and that models encode complementary rather than redundant
representations; however, concatenation of models is not always the solution to augment representation power.
Cross-referencing recoverability with per-species feature salience
(NMI) further revealed that task-relevant features are not always
well-encoded. These results suggest that
strategies compensating for weak representation points, such as
selecting models known to encode task-salient features, or
concatenating embeddings that cover complementary acoustic
properties, could directly improve downstream classification
performance in bioacoustics.
Our study has several limitations.
First, eGeMAPS features may be poorly extracted from signals, biasing ground truth labels. In this scenario, low $R^2$ values could indicate poor feature extraction, rather than poor feature encoding. Relatedly, the eGeMAPS feature set is optimized for human speech and may not capture the most relevant properties of non-human signals.
The systematically low F0 recoverability may partly reflect unreliable F0 extractors rather than absent encoding \parencite{best2025bioacoustic}.
This hypothesis motivates the need for bioacoustic-specific feature sets that can serve as ground truths in future evaluation studies.
Second, pooling embeddings across time can result in a loss of temporal information, especially with non-transformer based models and it may skew recoverability estimates of time-varying features.
Finally, our study did not include layer-wise analyses potentially overlooking layer-wise encoding dynamics described in speech-based models \parencite{cauzinille2025crossing, pasad2023comparative}.

We recommend future studies further investigate the relationship between feature decodability, feature importance, and also include model performance on downstream tasks. As it stands,
our framework demonstrates how recoverable features are, but does not yet indicate which features models rely on to perform classification. This would be the next step towards complete interpretable systems.
These axes of evaluation have been formalised in the evaluation framework proposed in \textcite{plachouras2025towards}.
Probing randomly initialized models would further help disentangle the role of architectural inductive biases from learned representations \parencite{ulyanov2018deep}, and combining our proposed framework with ablation or attention-based analyses could further clarify the link between embedding content and model behaviour \parencite{miron2026probing}.
Taken together, our findings contribute to a more interpretable and principled use of bioacoustic embeddings, moving beyond benchmarks to provide content-based guidance for model selection and expanding our understanding of what models have learned about the acoustic structure of non-human communication.




\clearpage
\printbibliography
\clearpage

\end{document}